\documentclass[conference]{IEEEtran}
\IEEEoverridecommandlockouts
\usepackage{graphicx} 
\usepackage{epsfig}
\usepackage{epstopdf} 
\usepackage{booktabs}
\usepackage{framed,multirow}
\usepackage{booktabs}

\usepackage{cite}
\usepackage{amsmath,amssymb,amsfonts}
\usepackage{algorithmic}
\usepackage{graphicx}
\usepackage{textcomp}
\usepackage{xcolor}

\usepackage{ragged2e} 
\usepackage{booktabs,makecell, multirow, tabularx}

\usepackage{amsmath,amssymb,amsfonts}
\usepackage{algorithmic}

\usepackage[numbers,sort&compress]{natbib}

\def\BibTeX{{\rm B\kern-.05em{\sc i\kern-.025em b}\kern-.08em
    T\kern-.1667em\lower.7ex\hbox{E}\kern-.125emX}}
\begin{document}

\title{Self-supervised Monocular Depth Estimation with Large Kernel Attention\\

\thanks{This work was supported in part by the National Natural Science Foundation of China under Grant  62271160 and 62176068, in part by the Natural Science Foundation of Heilongjiang Province of China under Grant LH2021F011, in part by the Fundamental Research Funds for the Central Universities of China under Grant 3072024LJ0803, in part by the Natural Science Foundation of Guangdong Province of China under Grant 2022A1515011527.}
}

\author{\IEEEauthorblockN{1\textsuperscript{st} Xuezhi Xiang}
	\IEEEauthorblockA{\textit{Harbin Engineering University} \\
		Harbin, China \\
		xiangxuezhi@hrbeu.edu.cn}
	
\and
\IEEEauthorblockN{2\textsuperscript{nd} Yao Wang}
\IEEEauthorblockA{\textit{Harbin Engineering University} \\
	Harbin, China \\
	wangyao2017080818@hrbeu.edu.cn}

\and
\IEEEauthorblockN{3\textsuperscript{rd} Lei Zhang}
\IEEEauthorblockA{\textit{Guangdong University of Petrochemical Technology} \\
	Maoming, China \\
	zhanglei@gdupt.edu.cn}

\and
\IEEEauthorblockN{4\textsuperscript{th} Denis Ombati}
\IEEEauthorblockA{\textit{Harbin Engineering University} \\
	Harbin, China \\
	lixiaohengy@hrbeu.edu.cn}

\and
\IEEEauthorblockN{5\textsuperscript{th} Himaloy Himu}
\IEEEauthorblockA{\textit{Harbin Engineering University} \\
	Harbin, China \\
	himaloy@hrbeu.edu.cn}
	
\and
\IEEEauthorblockN{6\textsuperscript{th} Xiantong Zhen}
\IEEEauthorblockA{\textit{Guangdong University of Petrochemical Technology} \\
	Maoming, China \\
	zhenxt@gmail.com}
}
\maketitle

\begin{abstract}
Self-supervised monocular depth estimation has emerged as a promising approach since it does not rely on labeled training data. Most methods combine convolution and Transformer to model long-distance dependencies to estimate depth accurately. However, Transformer treats 2D image features as 1D sequences, and positional encoding somewhat mitigates the loss of spatial information between different feature blocks, tending to overlook channel features, which limit the performance of depth estimation. In this paper, we propose a self-supervised monocular depth estimation network to get finer details. Specifically, we propose a decoder based on large kernel attention, which can model long-distance dependencies without compromising the two-dimension structure of features while maintaining feature channel adaptivity. In addition, we introduce a up-sampling module to accurately recover the fine details in the depth map. 
Our method achieves competitive results on the KITTI dataset.
\end{abstract}


\section{Introduction}
Monocular depth estimation is a fundamental computer vision task, aiming to estimate depth from single 2D image or video, and is widely used in autonomous driving, augmented reality and other fields. At present, monocular depth estimation based on deep learning \cite{ranftl2021vision, saxena2023monocular, ji2023ddp} has achieved excellent results. An inherent limitation of the supervised approach is that a large set of images with depth labels is required for training, while depth labels are expensive to acquire. Therefore, self-supervised monocular depth estimation \cite{godard2019digging, lyu2021hr, zhou2021self, he2022ra,han2023self, liu2024towards} has been recognized as a kind of promising approach. These methods use image reprojections from different viewpoints as supervision signals by exploiting geometric relationships between frames, i. e. scene depth and camera pose. However, view reconstruction loss is hindered by occlusions, dynamic objects, and photometric changes, which seriously affect the performance of the network. To address these challenges, researchers often incorporate novel constraints and utilize additional cues, such as semantic segmentation \cite{klingner2020self} and optical flow \cite{yin2018geonet}. Recently, self-supervised monocular depth estimation based on CNNs, Transformer and their variants \cite{zhou2021self, lyu2021hr, he2022ra, zhao2022monovit,bae2023deep} have achieved remarkable results, but Transformer and convolution still have their shortcomings. 

Previous methods \cite{zhao2022monovit, bae2023deep, han2022transdssl} use Transformer \cite{dosovitskiy2021image} to capture long-distance dependencies. However, self-attention treats 2D images as 1D sequences, which destroys the key 2D structure of images. Processing high-resolution images is also difficult due to its quadratic computation complexity and memory overhead. Moreover, Transformer only considers spatial dimension adaptation and ignores channel dimension adaptation. These limitations cause Transformer methods suffer from high computational cost and poor perception of small details as they focus more on long-distance information. Large kernel attention (LKA) \cite{guo2023visual}, which is tailored for vision tasks, can perfectly solve the above problems. We introduce it into monocular depth estimation, absorbing the advantages of convolution and self-attention, including local structural information, long-distance dependencies, and adaptability while avoids their shortcomings such as ignoring adaptivity in the channel dimension. By applying LKA to our depth network, we can improve the ability of the model to produce fine-grained and detailed depth map, avoiding blurring between foreground and background.  

Besides, a high-quality upsampler for self-supervised depth estimation should simultaneously recover the details, maintain the consistency of the depth value in a plain region, and also tackles gradually changed depth values. Previous methods used simple bilinear interpolation to recover the image in decoder, which often cause the blurred edges in feature maps. Inspired by \cite{guo2023visual, liu2023learning}, in this paper, we apply an upsample module in depth network to recover the fine depth and improve the accuracy of monocular depth estimation.

The contributions of this paper can be summarized as follows:
\begin{itemize}
\item We propose a self-supervised monocular depth network based on large kernel attention to improve the performance of depth estimation, which can model long-distance dependencies, while maintaining feature channel adaptivity without compromising the two-dimension structure of features, and improve estimation accuracy.
	
\item We introduce a upsample module to accurately recover the details in the depth map and improve the accuracy of monocular depth estimation. 
	
\item Extensive experiments demonstrate that our method achieves competitive performance on the KITTI dataset (AbsRel = 0.095, SqRel = 0.620, RMSE = 4.148, RMSElog = 0.169, $\delta$1 = 90.7).
\end{itemize}
\begin{figure}[h]
	\centering
	\includegraphics[width=0.5\textwidth]{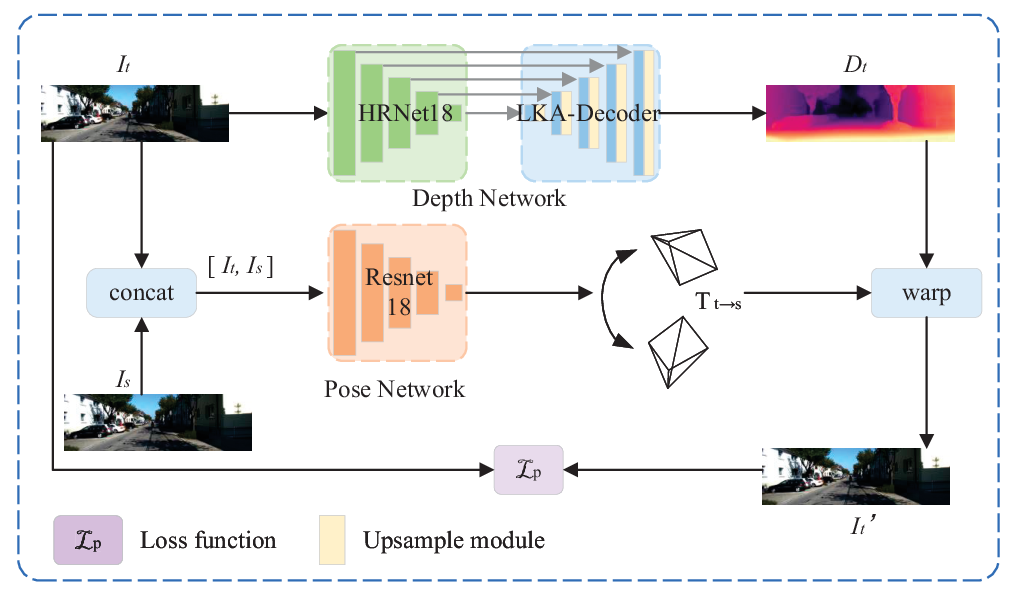}
	\caption{The overall architecture of our self-supervised monocular depth estimation method, which contains a depth network and a pose network. }
	\label{fig-overall}
\end{figure}

\section{Method}
\subsection{Overall architecture}\label{overall}
We take \cite{he2022ra} as baseline, and the overall architecture of our monocular depth estimation method is shown in Fig. \ref{fig-overall}, consisting of a depth network and a pose network. Our depth network adopts encoder-decoder architecture and HRNet18 \cite{lyu2021hr} is encoder, which provides multi-scale features by maintaining high-resolution representation through the entire process and repeatedly fusing the representation. The LKA-based decoder receives the features from the encoder. And we use ResNet18 as pose network to generate 6-DoF relative pose. 

\begin{figure}[h]
	\centering
	\includegraphics[width=0.49\textwidth]{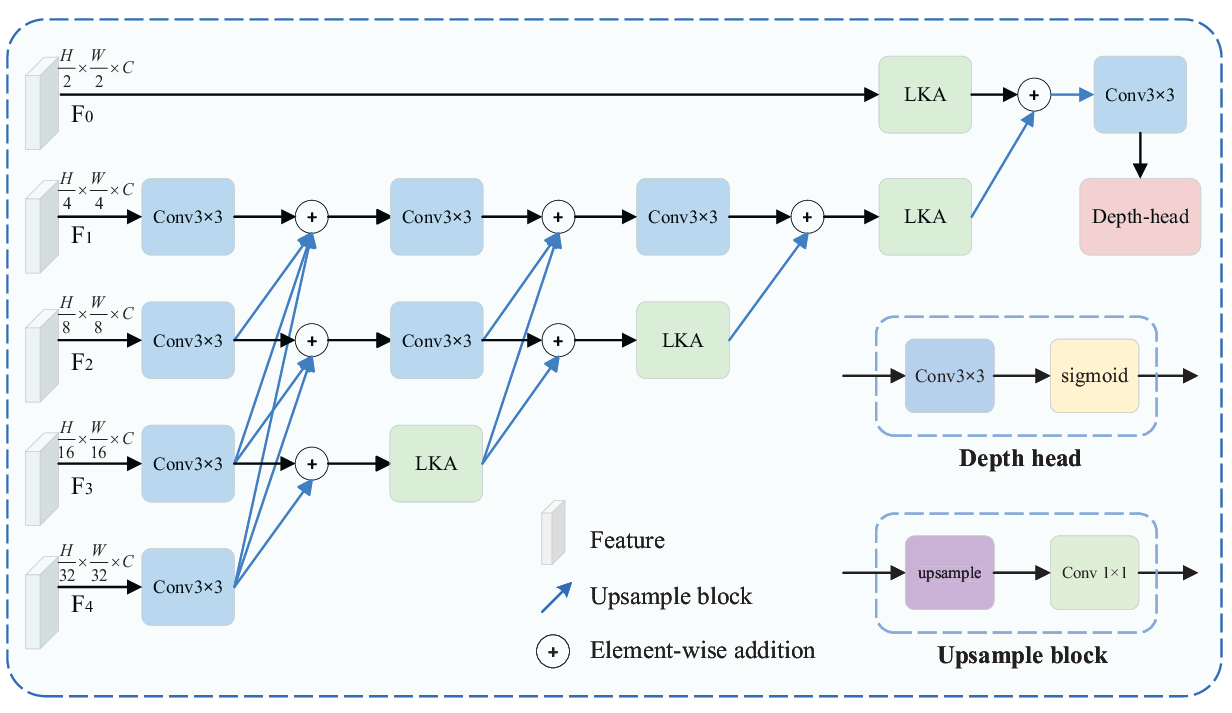}
	\caption{Overview of our depth decoder. The features are fed into ${3\times3}$ convolution layer and are concatenated with the upsampled features of the next layer and fed into LKA.}
	\label{fig-decoder}
\end{figure}

The proposed decoder applies LKA and upsampling moudle in every stage, as shown in Fig. \ref{fig-decoder}. The proposed decoder inherits the multi-scale features from the encoder and fuses lower-scale features while preserving high-resolution feature representations. Specifically, the features given by the encoder are fed into ${3\times3}$ convolution layer and are concatenated with the next layer features after upsampling. The concatenated features are fed into LKA and the final output is disparity map.

\subsection{Large kernel attention}\label{lka} 
\begin{figure}[h]
	\centering
	\includegraphics[width=0.49\textwidth]{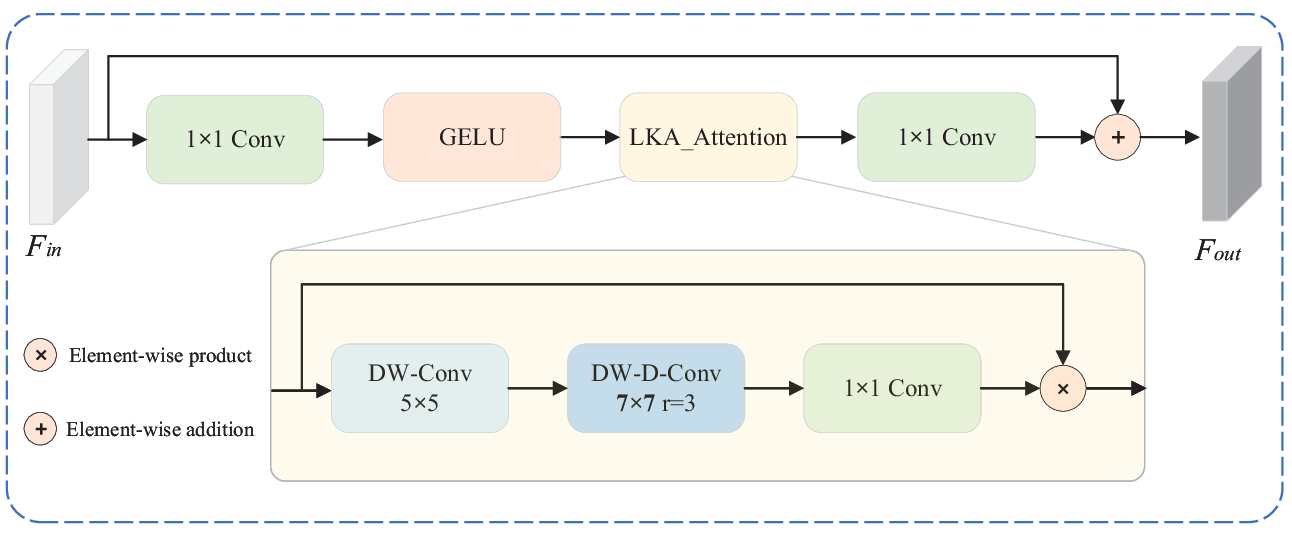}
	\caption{The architecture of large kernel attention (LKA). It is composed with depthwise convolution, depth-wise dilation convolution, and ${1\times1}$ convolution.}
	\label{fig-lka}
\end{figure}

The architecture of LKA is illustrated in Fig. \ref{fig-lka}, composed with a spatial local convolution (depthwise convolution), a spatial long-range convolution (depth-wise dilation convolution), and a channel convolution (${1\times1}$ convolution). Long-distance dependencies are modeled through cascaded depthwise separable convolutions and context features are obtained by a large kernel convolution, producing a feature with self-similarity in the appearance feature space. Subsequently, correlation is established through dot product operation, and it can be illustrated as: 
\begin{equation}\label{equa2}
	\emph{Attention}=\emph{Conv}_{1\times1}(\emph{DW-D-Conv}(\emph{DW-Conv}(\emph{F}_{in}))), 
	\tag{2}
\end{equation}
\begin{equation}\label{equa3}
	\emph{F}_{out}=Attention \otimes F_{in}.
	\tag{3}
\end{equation}
Through this decomposition, contextual information is recursively aggregated within the receptive field, gradually expanding effective receptive field. Larger receptive fields enable the proposed network to capture finer and more informative features, resulting the depth of scene can be estimated more accurately.

As a result, the proposed network can model long-distance dependencies while maintaining feature channel adaptivity without compromising the two-dimension structure of features, and improve depth estimation accuracy with lower computational cost and parameters. 

\subsection{Upsample module}
\begin{figure}[h]
	\centering
	\includegraphics[width=0.49\textwidth]{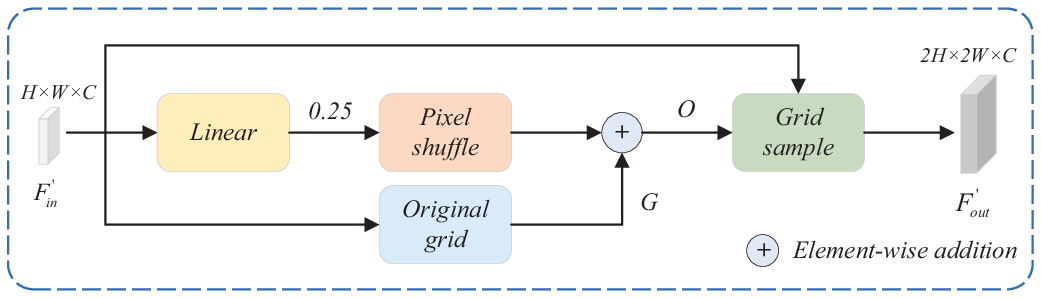}
	\caption{The architecture of upsample module. The grid sample function uses the offset to resample ${F}^{'}_{in}$ to ${F}^{'}_{out}$.}
	\label{fig-sample}
\end{figure}

A high-quality upsampler for self-supervised depth estimation should simultaneously recover the details, maintain the consistency of the depth value in a plain region, and also tackles gradually changed depth values. Previous methods used simple bilinear interpolation to recover the feature in decoder, which often cause the blurred edges in depth map and influence the prediction near boundaries. Such errors would propagate stage by stage, resulting in an unclear depth map. 
In our network, we introduce an upsample module to better estimate the depth, as shown in Fig. \ref{fig-sample}. Specifically, given the input feature ${F}^{'}_{in}$$\in R^{C\times H\times W}$, the offset \emph{O}$\in R^{2\times 2H\times 2W}$ is generated by linear layer and pixel shuffle \cite{shi2016realtime} then added to the original sampling grid. The grid sample function uses the offset positions to resample to ${F}^{'}_{out}$$\in R^{C\times 2H\times 2W}$, and it can be formulated as:
\begin{equation}\label{equa4}
	\emph{O}=\emph{PixelShuffle}(\emph{0.25$\times$Linear}(\emph{F}^{'}_{in})) + \emph{G},
	\tag{4}
\end{equation}
\begin{equation}\label{equa5}
	\emph{F}^{'}_{out}=\emph{GridSample}(\emph{F}^{'}_{in},\emph{O}),
	\tag{5}
\end{equation}
where $\emph{G}$ is the original sampling grid. By applying our upsample module instead of bilinear interpolation, the proposed decoder can recover the feature details more accurately.

\section{Experiments}
\subsection{Datasets and Metrics}
Our models are trained and evaluated on the KITTI datasets, adopting the data split of \cite{eigen2014depth} and follow pre-processing operation in \cite{zhou2017unsupervised} to remove static frames for training and testing. Finally, 39,810 frames are used for training, 4,424 for evaluation, and 697 frames for testing. And in the experimental evaluation process, the predicted depth is fixed in the range of 0 to 80m, as is common practice. 
We use seven commonly used metrics to evaluate our model, following \cite{zhou2017unsupervised}. For error metrics AbsRel, SqRel, RMSE and RMSElog, lower is better. For accuracy metrics ${\delta}<{1.25}$, ${\delta}<{1.25^2}$ , ${\delta}<{1.25^3}$ , higher is better.

\subsection{Implementation details}
The proposed model is implemented in PyTorch and trained on a single NVIDIA TITAN RTX GPU, with the batch size of 12. The initial learning rate is set to 1.0e-4 and decays to 1.0e-5 after 15 epochs with 20 epochs total. 

To make the network converge fast, we initialize HRNet18 and ResNet18 with the weights pretrained on ImageNet, and the input image resolution is uniformly cropped to a size of $640{\times}192$ during training to ensure the consistency of the experiments.

\begin{table*}[h]
	\centering
	\caption{Quantitative results on the KITTI dataset}
	\label{tab-kitti}
	\resizebox{1.0\linewidth}{!}{
		\begin{tabular}{l|c|c|cccc|ccc}
			\hline
		    \toprule[1pt] 
			\multicolumn{1}{c|}{Method} & Train & Resolution & AbsRel$\downarrow$ & SqRel$\downarrow$ & RMSE$\downarrow$  & RMSElog$\downarrow$ & $\delta<1.25$$\uparrow$ & $\delta<1.25^2$$\uparrow$ & $\delta<1.25^3$$\uparrow$    \\ \hline	
			SfMLearner \cite{zhou2017unsupervised}       & M     & 640×192    & 0.183  & 1.595  & 6.709  & 0.270   & 0.734            & 0.902             & 0.959    \\				
			Monodepth2 \cite{godard2019digging}          & M     & 640×192    & 0.115  & 0.903  & 4.863  & 0.193   & 0.877            & 0.959             & 0.981      \\
			SGDDepth \cite{klingner2020self}             & M+Se  & 640×192    & 0.113  & 0.835  & 4.693  & 0.191   & 0.879            & 0.961             & 0.981     \\
			SAFENet \cite{choi2020safenet}               & M+Se  & 640×192    & 0.112  & 0.788  & 4.582  & 0.187   & 0.878            & 0.963             & 0.983      \\
			PackNet-SfM \cite{guizilini20203d}           & M     & 640×192    & 0.111  & 0.785  & 4.601  & 0.189   & 0.878            & 0.960             & 0.982      \\
			Mono-certainty\cite{poggi2020uncertainty}    & M     & 640×192    & 0.111  & 0.863  & 4.756  & 0.188   & 0.881            & 0.961             & 0.982             \\
			HR-Depth \cite{lyu2021hr}                    & M     & 640×192    & 0.109  & 0.792  & 4.632  & 0.185   & 0.884            & 0.962             & 0.983       \\
			DIFFNet \cite{zhou2021self}                  & M     & 640×192    & 0.102  & 0.764  & 4.483  & 0.180   & 0.896            & 0.965             & 0.983        \\
			CADepth \cite{yan2021channelwise}            & M     & 640×192    & 0.105  & 0.769  & 4.535  & 0.181   & 0.892            & 0.964             & 0.983      \\
			TransDSSL \cite{han2022transdssl}            & M     & 640×192    & 0.098  & 0.728  & 4.458  & 0.176   & 0.898            & 0.966             & 0.984      \\
			MonoViT \cite{zhao2022monovit}               & M     & 640×192    & 0.099  & 0.708  & 4.372  & 0.175   & 0.900            & 0.967             & 0.984     \\
			RA-Depth \cite{he2022ra}                     & M     & 640×192    & \underline{0.096}  & 0.632  & 4.216  & 0.171  & 0.903 & 0.968             & \underline{0.985} \\
			MonoFormer \cite{bae2023deep}                & M     & 640×192    & 0.104  & 0.846  & 4.580  & 0.183   & 0.891            & 0.962             & 0.982              \\
			DaCCN \cite{han2023self}                     & M     & 640×192    & 0.099  & 0.661  & 4.316  & 0.173   & 0.897            & 0.967             & \underline{0.985}   \\
			MonoVan \cite{indky2023monovan}              & M     & 640×192    & 0.101  & 0.706  & 4.416  & 0.176   & 0.897            & 0.966             & 0.984           \\
			BDEdepth \cite{liu2024towards}             & M  & 640×192  & \textbf{0.095}  & \underline{0.621}  & \underline{4.183}   & \underline{0.170}   & \underline{0.904}  & 0.968   & \underline{0.985}   \\
			MambaDepth \cite{grigore2024mambade}       & M  & 640×192  & 0.097  & 0.706  & 4.370  & 0.172   & \textbf{0.907}         & \textbf{0.970}           & \textbf{0.986}     \\
			Ours                                       & M  & 640×192  & \textbf{0.095}  & \textbf{0.620} & \textbf{4.148} & \textbf{0.169} & \textbf{0.907} & \underline{0.969}  & \underline{0.985}   \\ 
		    \bottomrule[1pt]
			\hline
			 
			\multicolumn{10}{p{1.8\columnwidth}}{Comparison of our method to existing methods on the KITTI dataset using the Eigen split. M: trained with monocular videos; Se: Trained with semantic labels. The best results in each category are in \textbf{bold} and the second best are \underline{underlined}.}
		\end{tabular}
	}
\end{table*}

\begin{table*}[h]
	\centering
	\caption{Ablation results for each component of our method on the KITTI dataset}
	\label{tab-ablation}
	\resizebox{1.0\linewidth}{!} {  %wy: 
		\begin{tabular}{l|cc|cccc|ccc|cc}
			\hline
			\toprule[1pt]
			\multicolumn{1}{c|}{Method} & LKA & upsample & AbsRel$\downarrow$ & SqRel$\downarrow$ & RMSE$\downarrow$ & RMSElog$\downarrow$ & $\delta<1.25$$\uparrow$ & $\delta<1.25^2$$\uparrow$ & $\delta<1.25^3$$\uparrow$   & Params (M)  & GFLOPs    \\ \hline
			Baseline                   &     &            & 0.096          & 0.632         & 4.216         & 0.171       & 0.903         & 0.968      & 0.985    & 9.98   & 10.78  \\ 
			\multirow{3}{*}{Ours}      & \checkmark   &   & 0.095          &  \textbf{0.617}         & 4.168         & 0.170       & 0.905         & 0.969      & 0.985    & 9.93   & 10.16\\                                                
		                           	&             & \checkmark  & 0.096          & 0.621        & 4.165         & 0.170       & 0.905         & 0.968      & 0.985    & 9.98   & 10.83  \\                         
		                        	& \checkmark      &  \checkmark           &  \textbf{0.095}  & 0.620  &  \textbf{4.148}  &  \textbf{0.169}  &  \textbf{0.907}  &  \textbf{0.969}  &  \textbf{0.985}  & 9.93  & 10.19 \\  
	\bottomrule[1pt] 
	\hline	
			
			\multicolumn{12}{p{1.9\columnwidth}}{LKA: large kernel attention. The best results are in \textbf{bold}.}
		\end{tabular}
}
\end{table*}

\subsection{Results}

\textbf{Quantitative results}. The experimental results on KITTI dataset are presented in Table \ref{tab-kitti}, and all models are tested at the same resolution (640$\times$192) as training.  Benefit from the proposed decoder, our method can model long-distance dependencies between pixels, and capture more accurate context information to process complex scene. As a result, our method shows higher estimation accuracy ($\delta$1, $\delta$2, $\delta$3) and lower error (AbsRel, SqRel, RMSE, RMSElog). Specifically, compared with Transformer methods MonoVit \cite{zhao2022monovit} and MonoFormer \cite{bae2023deep}, our method outperforms on all metrics (with AbsRel, SqRel, RMSE and RMSElog decreasing by 8.7$\%$, 26.7$\%$, 9.4$\%$ and 7.7$\%$, respectively and $\delta1$ increasing by 1.8$\%$). Compared with CNN methods HR-Depth \cite{lyu2021hr}, DIFFNet \cite{zhou2021self} and RA-Depth \cite{he2022ra}, our model also achieves superior performance. Besides, compared to the BDEdepth \cite{liu2024towards}, which apply a grid decoder to enhance details in depth map, the proposed method also achieve better performance with almost same parameters, which means our decoder performs better. It is worth to mention that compared with MonoVan \cite{indky2023monovan}, which use VAN as backbone, we also achieve superior performance with less parameters (with AbsRel, SqRel, RMSE and RMSElog decreasing by 5.9$\%$, 12.2$\%$, 6.1$\%$ and 4.0$\%$, respectively). 
In a word, compared with existing methods, our method shows superior performance and even achieves the best performance on some metrics (AbsRel = 0.095, SqRel = 0.620, RMSE = 4.148, RMSElog = 0.169, $\delta$1 = 90.7).

\begin{figure}[!h]
	\centering
	\includegraphics[width=0.5\textwidth]{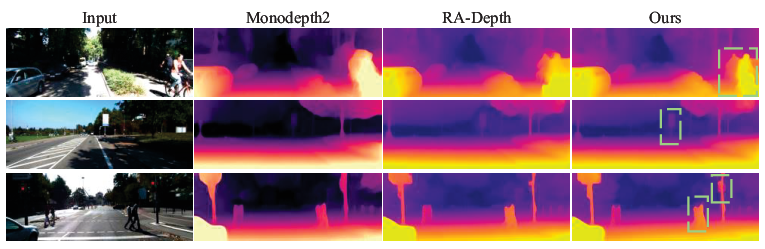}
	\caption{Qualitative results on the KITTI dataset.Our model can obtain higher quality depth maps with finer depth edges compared to other methods.}
	\label{fig-kitti}
\end{figure}

\textbf{Qualitative results}. We provide qualitative comparison results on different scenes of the KITTI dataset, comparing with our baseline \cite{he2022ra} and the classic work Monodepth2 \cite{godard2019digging}, as shown in Fig. \ref{fig-kitti}. Monodepth2 and RA-depth have limited receptive fields, so they yield some
inaccurate depth predictions. Instead, our models can generate better results. It can be seen that our method distinguish the boundaries (traffic signs, pedestrians and roadside trees) in the scene more clearly. As a result, we obtain higher quality depth maps with sharper depth edges. This is mainly benefit from that our network can capture more accurate spatial information, exhibit superior control over the foreground and background in the scene, resulting in more sharper edges and higher accuracy.

\subsection{Ablation study}

In this section, we conduct ablation experiments on the KITTI dataset to validate the effectiveness of the proposed method, and Table \ref{tab-ablation} shows the experiment results. 

\textbf{The benefit of large kernel attention}. We fistly employ the LKA in our decoder only. It can be seen that adding the LKA can improve the performance of depth estimation performance obviously. Specifically, compared to the baseline, the results show improvements especially in terms of SqRel and RMSE (with decreasing by 2.4$\%$ and 1.1$\%$ respectively), which indicates that the proposed decoder based on LKA enhances the accuracy of object boundaries’ depth predictions since most large depth errors occur at these boundaries. Furthermore, there is no additional parameter and computational consumption during inference. 

\textbf{The benefit of upsample}. We then employ the upsampling module in our decoder only. It can be seen that compared with the baseline, the error of our method is obviously decreased with no additional parameter, especially SqRel and RMSE (with decreasing by 1.7$\%$ and 1.2$\%$ respectively). This is mainly because upsample module accurately recover the details features and reduce the blurred edges in the depth map, as a result, the depth network can distinguish the boundary in the scene obviously, and then predict more exactly.

When two modules work together, in comparison with baseline, our method also demonstrates improvements, particularly in terms of RMSE and RMSElog (with decreasing by 1.6$\%$ and 1.2$\%$, respectively). Improvements in all metrics indicate that our method obtains a better performance for monocular depth estimation.

\textbf{Model efficiency}. Besides, our model demonstrates efficiency in terms of parameter and computation complexity. Specifically, our method shows excellent performance in error and accuracy with no addition in parameters compared with baseline, which means our method achieves a great balance between performance and efficiency.

\section{Conclusion}
\label{5-Conclusions}
In this paper, we propose a self-supervised monocular depth estimation network to get finer details and sharper edges. Specifically, we propose a depth decoder based on large kernel attention for self-supervised monocular depth estimation, which can model long-distance dependencies without compromising the two-dimension structure of features and improve estimation accuracy, while maintaining feature channel adaptivity. In addition, we introduce a up-sampling module, which can accurately recover the fine details in the depth map. Experiments demonstrate that our method exhibits excellent performance in predicting the depth of scene details. The proposed method achieves competitive results on the KITTI dataset.

\bibliographystyle{IEEEtran}
\bibliography{refs}

\end{document}